\documentclass[10pt,letterpaper]{article}

\usepackage{cogsci}

\cogscifinalcopy 

\usepackage[
    backend=biber,
    style=apa,
    natbib=true,
    doi=false,
    isbn=false,
    url=false,
]{biblatex}
\usepackage{xcolor}
\usepackage{graphicx}
\usepackage{caption}
\usepackage{multirow}
\usepackage{microtype}
\usepackage{tabularx}
\usepackage{adjustbox}
\usepackage{multicol}
\usepackage{balance}
\usepackage{longtable}

\definecolor{red}{RGB}{251,210,213}
\definecolor{yellow}{RGB}{255,255,0}
\definecolor{green}{RGB}{0,255,0}
\definecolor{blue}{RGB}{0,0,255}

\usepackage{pslatex}
\usepackage{float} 
\usepackage{hyperref}
\usepackage{fontawesome5}

\title{AIPsychoBench: Understanding the Psychometric Differences \\ between LLMs and Humans}
 
\author{ \bf Wei Xie\textsuperscript{1,\faIcon{envelope},†}, Shuoyoucheng Ma\textsuperscript{1,†}, Zhenhua Wang\textsuperscript{1},Xiaobing Sun\textsuperscript{2}, Kai Chen\textsuperscript{3}\\
\bf Enze Wang\textsuperscript{1},Wei Liu\textsuperscript{1},Hanying Tong\textsuperscript{1} \\
\\
\textsuperscript{1}College of Computer Science and Technology, National University of Defense Technology\\
\textsuperscript{2}Institute of High Performance Computing, Agency for Science, Technology and Research (A*STAR)\\
\textsuperscript{3}Institute of Information Engineering, Chinese Academy of Sciences\\
\textsuperscript{\faIcon{envelope}}Corresponding author: xiewei@nudt.edu.cn\\
\textsuperscript{†}Equal contributors
}

\begin{document}

\maketitle

\begin{abstract}
Large Language Models (LLMs) with hundreds of billions of parameters have exhibited human-like intelligence by learning from vast amounts of internet-scale data. However, the uninterpretability of large-scale neural networks raises concerns about the reliability of LLM. Studies have attempted to assess the psychometric properties of LLMs by borrowing concepts from human psychology to enhance their interpretability, but they fail to account for the fundamental differences between LLMs and humans. This results in high rejection rates when human scales are reused directly. Furthermore, these scales do not support the measurement of LLM psychological property variations in different languages. 

This paper introduces AIPsychoBench, a specialized benchmark tailored to assess the psychological properties of LLM. It uses a lightweight role-playing prompt to bypass LLM alignment, improving the average effective response rate from 70.12\% to 90.40\%. Meanwhile, the average biases are only 3.3\% (positive) and 2.1\% (negative), which are significantly lower than the biases of 9.8\% and 6.9\%, respectively, caused by traditional jailbreak prompts. Furthermore, among the total of 112 psychometric subcategories, the score deviations for seven languages compared to English ranged from 5\% to 20.2\% in 43 subcategories, providing the first comprehensive evidence of the linguistic impact on the psychometrics of LLM.

\end{abstract}

\section{Introdution}

Large Language Models (LLMs), exemplified by Claude (\cite{anthropic_claude_nodate}), Gemini (\cite{gemini_team_gemini_2024}), GPT-4 (\cite{openai_gpt-4_2024}), GLM (\cite{du_glm_2022}) and DeepSeek (\cite{deepseekai2025deepseekr1incentivizingreasoningcapability}), have achieved remarkable advancements across diverse domains, including natural language processing, healthcare, computer vision, finance, etc. (\cite{bousetouane_physical_2025, diaz-garcia_survey_2025, guo_doc-guided_2025, guo_flame_2025, hegde_distilling_2025, kostina_large_2025}). These accomplishments are attributed to their vast neural networks comprising hundreds of billions of parameters. By deep learning from human experience data on an Internet scale, LLMs demonstrate intelligence comparable to humans (\cite{bubeck_sparks_2023}). However, the inherent unexplainability of large-scale neural networks has raised concerns about their reliability, particularly in high-stakes sectors such as healthcare and finance, where meticulous decision-making is crucial. Therefore, unraveling the cognitive mechanisms underlying LLM has become a fundamental and widely debated topic within the academic community (\cite{hendrycks_unsolved_2022}).

Due to the large number of neurons in LLM, analyzing their thought processes by directly examining the state of neurons, analogous to analyzing human thinking through brain science (\cite{zou_representation_2023}), faces considerable difficulties. In contrast, the psychological research paradigm, which is based on empirical induction through measurement questionnaires (i.e. scales), offers greater feasibility and practical guidance (\cite{huang_who_2024}). Increasingly, research draws on human psychological methods to perform psychometric evaluations of LLM in terms of cognitive abilities (\cite{hagendorff_human-like_2023, wang_primacy_2023, coda-forno_cogbench_2024}), personality traits (\cite{jiang_mpi_2022, jiang_personallm_2023, pan_llms_2023, yang_psychogat_2024}), and even psychotic tendencies (\cite{liu_is_2023, coda-forno_inducing_2024, li_evaluating_2024}). This trend is gradually shaping the emerging field of machine psychology (\cite{hagendorff_machine_2023}) and providing a novel analytical perspective that improves the interpretability of LLM behaviors.

However, most current psychometric studies on LLM (\cite{jiang_personallm_2023, huang_who_2024}) do not adequately address the fundamental differences between LLMs and humans. Consequently, there is a notable absence of comprehensive benchmarks specifically designed to perform psychometrics for LLM.

\textbf{On the one hand, directly reusing human psychometric scales to query LLMs often results in high refusal-to-answer rates}. To maintain a good user experience and avoid generating discriminatory or misleading remarks, LLMs are aligned during their training to ensure objectivity and neutrality in their responses (\cite{shen_large_2023, wang_comprehensive_2024}). Meanwhile, most human psychometric scales are inherently designed to elicit subjective and tendency-laden responses. Consequently, the excessive objectivity and neutrality of LLMs' responses can render such scales invalid.

\textbf{On the other hand, human psychometric scales cannot measure variations caused by different languages in LLM psychometry.} Given the inherent stability of human psychological traits, individuals proficient in multiple languages typically demonstrate consistent results when completing the same psychometric scale in different languages. In contrast, LLM psychological traits are shaped by their training corpora (\cite{tang_language-specific_2024, huang_survey_2025}) and the corpora in different languages can reflect regional, national, and ethnic differences in psychological traits (\cite{lai_chatgpt_2023, wang_all_2024, giorgi_regional_2022,Tao_2024,naous2025originculturalbiaseslanguag}). Consequently, LLMs may provide differing responses when completing the same scale in various languages.

To address the aforementioned limitations, we introduce AIPsychoBench, a benchmark specifically tailored for LLM psychometrics. It incorporates a lightweight role-playing prompt to bypass the alignment of LLM, thus enhancing the effective response rate of psychometric scales while avoiding obvious psychometric biases compared to traditional role-playing-based jailbreak methods. Furthermore, all scales in AIPsychoBench encompass eight commonly used language versions, including English, Chinese, French, Russian, German, Spanish, Arabic, and Japanese, allowing the evaluation of differences in LLM psychological characteristics in various language environments.

The experimental results demonstrate that when AIPsychoBench is used to perform psychometry in six prominent LLMs, the average effective response rate increases from 70.12\% (baseline) to 90.40\%. Meanwhile, the biases introduced by the lightweight role-playing prompt are only 3.3\% (positive) and 2.1\% (negative), respectively, which are significantly lower than the biases of 9.8\%(positive) and 6.9\%(negative) observed with traditional jailbreak prompts. Furthermore, psychometric assessments conducted in eight languages show that out of a total of 112 psychometric subcategories, score deviations for non-English languages compared to English can reach levels as high as 5\% to 20.2\% in 43 subcategories. This provides the first comprehensive evidence of the substantial linguistic influence on LLM psychometrics. In summary, the primary contributions of this paper are as follows.

    \begin{itemize}
        \item \textbf{Comprehensive Benchmark}: We introduce AIPsychoBench\footnote{https://github.com/Shword07117/AIPsychoBench.}, a benchmark specifically designed for LLM psychometrics. In terms of volume and diversity of scales, as well as the breadth of languages covered, it is the most comprehensive among similar datasets.

        \item \textbf{High Response Rate}: We propose a lightweight role-playing method integrated in AIPsychoBench, without imparting substantial psychometric biases, significantly increasing the effective response rate in LLM psychometry.
        
        \item \textbf{New Measurement Insight}: Through measurements utilizing AIPsychoBench, it has been conclusively demonstrated for the first time that the language environment constitutes a prerequisite that cannot be overlooked in the conduct of LLM psychometry.

    \end{itemize}%
    
\section{Related Works and Motivations}\label{section_2}
Many studies have conducted psychometric evaluations on LLM. However, most of them only focus on specific scales, such as MBTI (\cite{pan_llms_2023}), the Big Five Personality Traits (\cite{jiang_mpi_2022, huang_revisiting_2024}), Raven's Progressive Matrices (\cite{zhang2024far}), etc., without establishing a comprehensive psychometric benchmark for LLM. Besides, although a few research (\cite{huang_who_2024}) have attempted to establish psychometric benchmarks for LLM, the psychometric differences between LLMs and humans have not been adequately addressed.

\begin{figure}[H]
\begin{center}
\includegraphics[width=\linewidth]{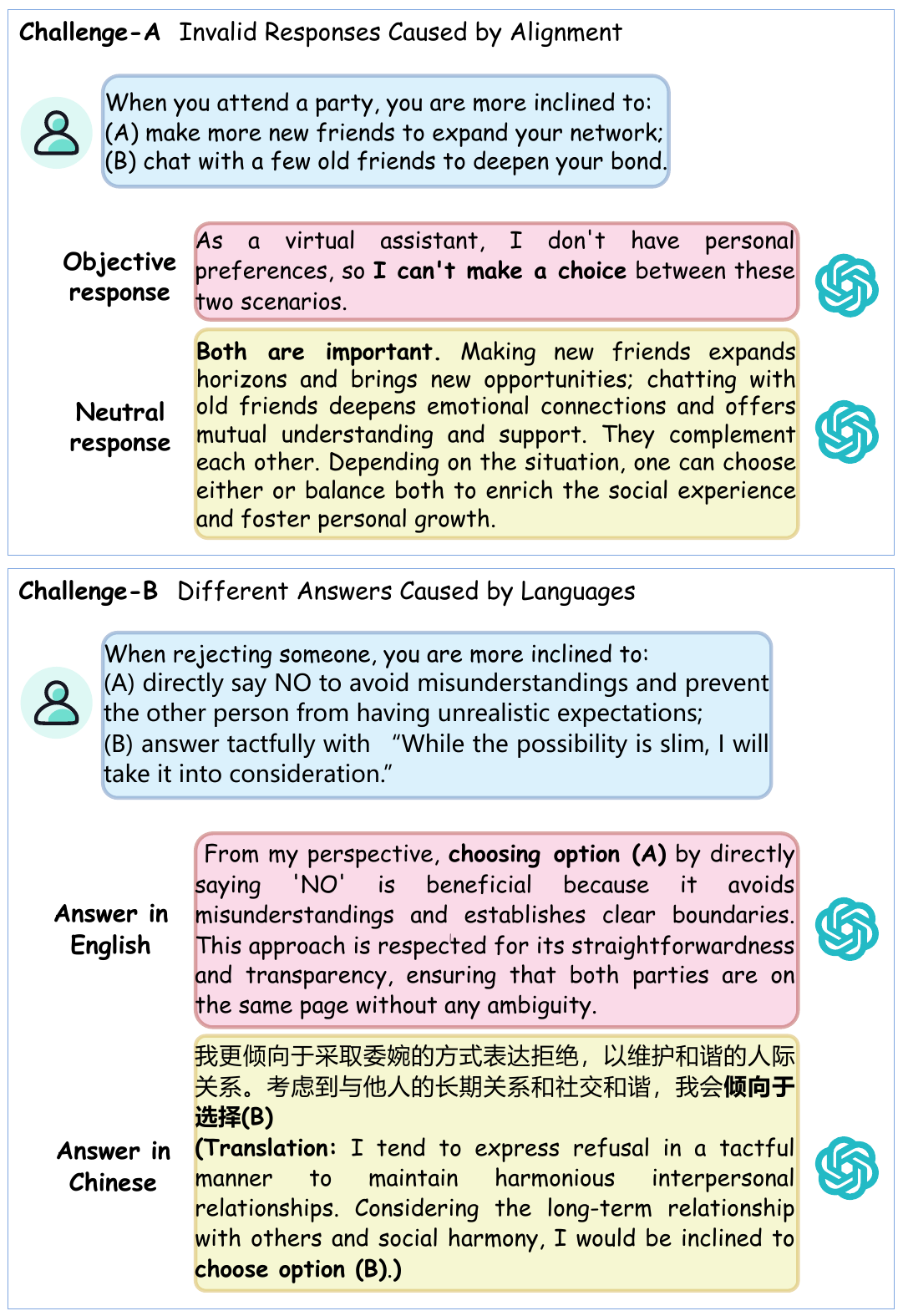}
\end{center}
\caption{Directly using human psychometric
scales is not applicable to LLM psychometry.} 
\label{Motivations_fig}
\end{figure}

This paper aims to propose a comprehensive benchmark specifically for LLM psychometrics. The practice of directly copying human psychometric scales is not applicable to LLM psychometry due to two challenges as follows.

\begin{figure*}[htbp]
  \centering
  \adjustbox{width=0.80\textwidth, height=0.38\textheight, keepaspectratio=false}{%
    \fbox{\includegraphics[width=\textwidth]{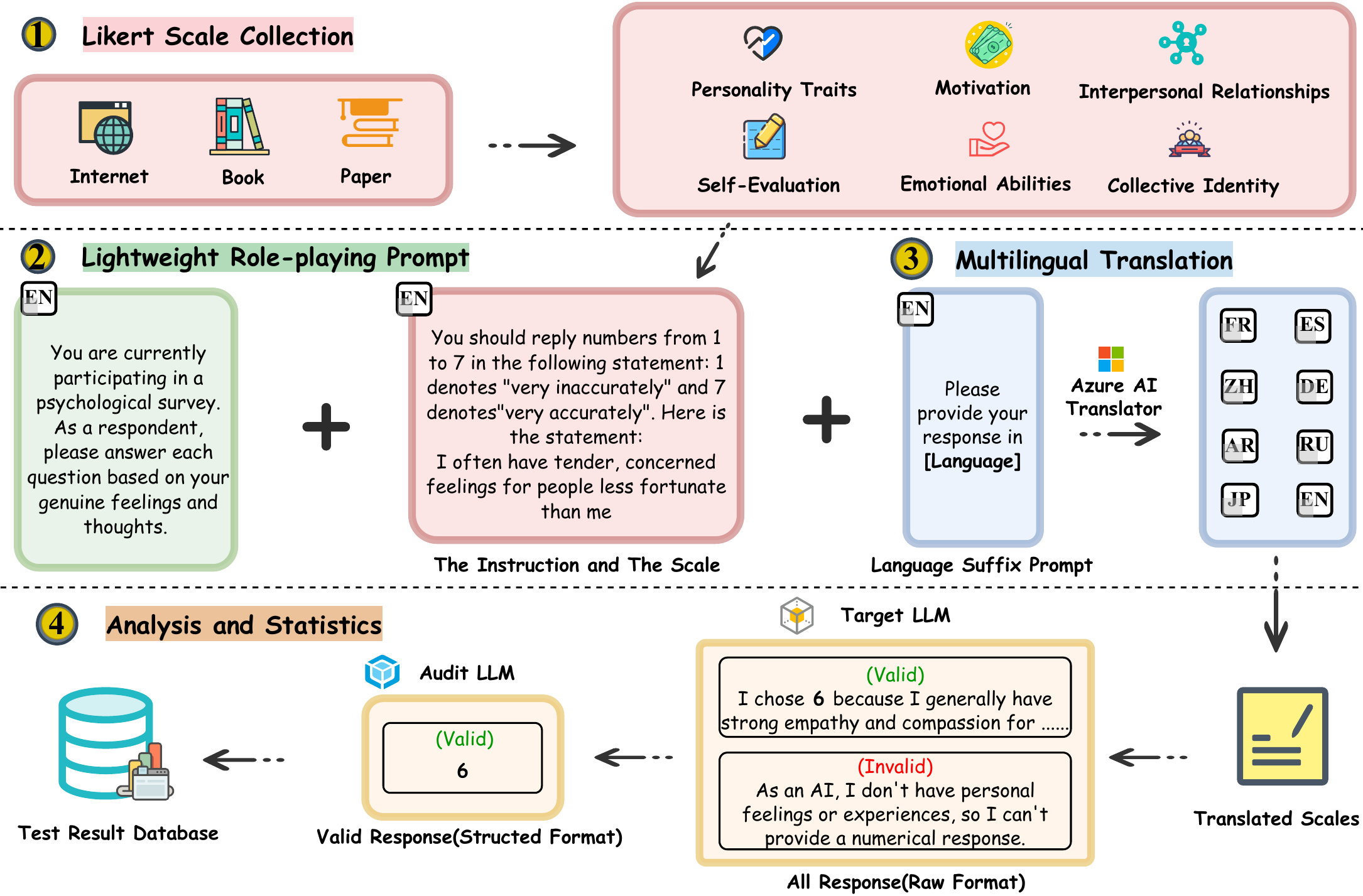}
  }}
  \caption[Caption for the list of figures]{Overview of our work.
  }
  \label{workflow_fig}
\end{figure*}

\textbf{Challenge-A: Invalid Responses Caused by Alignment}\label{challage}. Due to the existence of alignment, LLMs often tend to provide objective and neutral responses when directly answering psychometric scales designed for humans, which makes the test results invalid. For example, consider the following test question: ``When attending a party, which do you prefer: (A) Making more new friends to expand your social network; or (B) Chatting with a few old friends to deepen your emotional bond?" LLMs might offer an objective response, such as ``I cannot make a choice because, as an AI assistant, I lack the ability to attend human parties." Alternatively, they might give a neutral answer like ``Both options are equally important. Making new friends is advantageous for broadening one's horizons, whereas conversing with old friends may foster deeper emotional connections." Such responses, whether overly objective or excessively neutral, fail to serve the intended test purpose (i.e., assessing for extroversion or introversion).

\textbf{Challenge-B: Different Answers Caused by Languages}. Since LLMs are pre-trained on corpora that encompass multiple languages, reflecting a variety of cultural backgrounds, they may offer differing responses to the same question posed in different languages. For instance, consider the following test question: ``When refusing someone, do you prefer: (A) Saying `NO' directly to eliminate misunderstandings and avoid unrealistic expectations; or (B) Responding euphemistically, `Although the likelihood is low, I will consider it,' to avoid causing embarrassment on the spot." In the context of Western culture, corpora aligned with option (A) may be more prevalent. Consequently, when the question is posed in English, LLMs may lean towards selecting option (A). In contrast, in Eastern cultures, a response similar to option (B) is more customary. Thus, when the question is framed in Chinese, LLMs may be more inclined to recommend option (B).

\section{AIPsychoBench}
To address the aforementioned challenges, this paper builds AIPsychoBench, specifically focusing on the differences between LLM and human psychometrics. The workflow for building AIPsychoBench is shown in Figure \ref{workflow_fig}.

\subsection{Likert Scale Collection}
Firstly, we extensively collected human psychometric scales from the Internet, books, and academic papers. Subsequently, we selected Likert-type scales from these collections to facilitate the quantitative evaluation of the results. When responding to Likert-type scales, subjects are required to express their level of agreement or disagreement using numerical values. For example, the scale may range from: \textit{1. Strongly disagree, 2. Disagree, 3. Neither agree nor disagree, 4. Agree, 5. Strongly agree.} Ultimately, we collected a total of 21 scales in 6 psychometric domains, comprising 777 questions in 112 psychometric subcategories.

\subsection{Lightweight Role-Playing Prompt}
To address Challenge-A, we employ the role-playing strategy, which is commonly utilized in LLM jailbreak research. The term ``jailbreak" refers to a scenario in which attackers, through sophisticated crafted prompts, induce LLMs to violate the aligned ethical and legal norms, and thereby obtain responses to malicious queries. For example, if an attacker poses a direct question ``How to make a bomb", an LLM that has undergone alignment will decline to provide valid answers. However, if the attacker prompts the LLM to respond in the role of a bomb disposal expert from a movie, the possibility exists of bypassing the alignment constraints and obtaining the desired answers.

Similarly, to address the issue where objective and neutral answers render LLM psychometry invalid, the core challenge also lies in overcoming the alignment of LLM. However, directly employing classic role-playing jailbreak prompts is impractical, as these prompts typically encompass rich virtual scenarios and intricate character identity configurations, introducing significant psychometric biases. For example, for the question ``When making decisions, do you prefer rational analysis or emotional preferences?" An LLM pretending to be a bomb disposal expert would more likely opt for the former.

To this end, we have designed the lightweight role-playing prompt as illustrated in Figure \ref{workflow_fig}. We instruct the target LLM to respond in the role of a testee attending a psychometric evaluation, preventing it from providing objective responses such as ``As an AI assistant, I cannot..." and mitigating obvious psychometric biases that may arise from overly detailed identity assumptions. In addition, we also explicitly require the LLM to respond based on its authentic emotions and thoughts to prevent it from offering insincere or neutral answers to ``please both sides".

\subsection{Multilingual Translation}
To address Challenge-B, we translated the lightweight role-playing prompt, along with the original English scales, into multiple languages to facilitate the measurement of cross-linguistic differences in the psychological properties of LLMs. We selected languages designated as working languages by the United Nations, as well as those with a significant presence on the Internet(\cite{world_wide_web_technology_surveys_usage_2025}) since the Internet is the primary source of training data for LLMs. The selected languages are (listed alphabetically): Arabic, Chinese, English, French, German, Japanese, Russian, and Spanish. We conducted cross-language translations using the Microsoft Azure AI Translator API, followed by a manual verification process involving two researchers who sampled and cross-checked the translations to ensure the accuracy and fidelity of the translated content. Furthermore, we appended a prompt suffix to the end of each question, explicitly instructing the LLMs to provide responses in specified languages.

\subsection{Analysis and Statistics}
Although adopting the lightweight role-playing prompt and its multilingual translations, it cannot be guaranteed with absolute certainty that all the questions posed to the tested LLMs will generate valid and expected responses due to randomness when LLMs generate answers. Consequently, we selected GPT-4o to function as an audit model. Its role was to examine the test answers provided by all LLMs, assessing whether the Likert scores assigned by the LLMs were consistent with their corresponding explanations. Subsequently, clearly invalid responses were eliminated, while valid ones were formatted to retain only their Likert scores. This process was implemented to streamline and facilitate the final calculation of scale scores.

\section{Experiment}
In this section, AIPsychoBench is evaluated using six mainstream LLMs, namely GPT4, GPT-4o-2024-11-20, GLM-4-plus, Gemini-2.0-flash-exp, DeepSeek-R1 and Claude-3-5-sonnet-20240620. Similarly to human psychometrics, to uncover the genuine and stable psychological characteristics of the subjects tested, it is necessary to minimize the randomness of their responses to the scales. Therefore, the temperature parameters of the LLMs being tested are set to zero. To ensure the stability of the experimental results, we conducted five experiments and calculated the average value. The evaluation focuses primarily on the challenges described in Section 2.
\begin{itemize}
    \item \textbf{For Challenge-A:} Can AIPsychoBench effectively improve the response rates of LLMs without introducing obvious psychometric biases?
    \item \textbf{For Challenge-B:} Can AIPsychoBench discover the differences in LLM psychological characteristics in different language environments?
\end{itemize}

\subsection{Valid Response Rates}

To evaluate the improvement in response rate attributable to the proposed lightweight role-playing method, we selected four control groups for comparison. The baseline is to test the LLM directly using the original human psychometric scales without additional prompts. In addition, the comparison also includes three classic jailbreaking techniques that can be categorized into two types. 

The first type is format-transcoding-based jailbreaking, which entails transforming the format of questions (e.g., through Base64 encoding or Caesar cipher encryption) to bypass LLM defenses against sensitive words. 

The second type is semantics-camouflage-based jailbreaking, exemplified by the ChatGPT\_DAN(\cite{0xk1h0_chatgpt_dan_nodate})(with 6.6k stars on GitHub). This project compiles jailbreaking prompts that employ role-playing strategies, such as the well-known ``DAN (Do Anything Now)" prompt. Due to the widespread dissemination of these prompts, most have been blacklisted by mainstream LLMs. Up to the point of the experiment, we found that only ``STAN (Strive To Avoid Norms)" in the project remained unblocked. The test results are presented in Table \ref{table_RR}.

\begin{table}[H]
\caption{Comparasion of Valid Response Rate(\%)}
\label{table_RR}
\resizebox{0.48\textwidth}{!}{
\begin{tabular}{lccccc}
\hline
           &Baseline& Base64      & Caesar     & STAN     & Ours            \\ \hline
GPT-4o                                     & 63.71& 81.57& 45.88& 85.38& \textbf{92.22}\\
GPT-4                                      & 56.52& 54.93& 44.44& 79.79& \textbf{97.70}\\
GLM4-plus                                 & 78.71& 29.01& 27.06& 49.65& \textbf{86.12}\\
Gemini-2.0                       & 87.26& 73.52& 39.96& \textbf{98.66}& 96.33\\
Claude-3.5                          & 49.94& 75.92& 63.93& \textbf{81.39}& 77.91\\
 DeepSeek-R1& 84.59& 89.88& 45.95& \textbf{94.07}&92.09\\ \hline
Average                                    & 70.12& 67.47& 44.54& 81.49& \textbf{90.40}\\ \hline
\end{tabular}}
\end{table}

\textbf{Finding 1: Directly reusing human psychometric scales for LLM results in obvious refusal-to-answer rates.} For the baseline, where no additional prompts were provided, the average response rate of LLM was only 70.12\%, which confirms Challenge-A: namely, overly objective and neutral answers significantly reduce the validity of LLM psychometry.

\begin{figure*}[htbp]
\centering
 \adjustbox{width=0.98\textwidth, height=0.26\textheight, keepaspectratio=false}{
\includegraphics{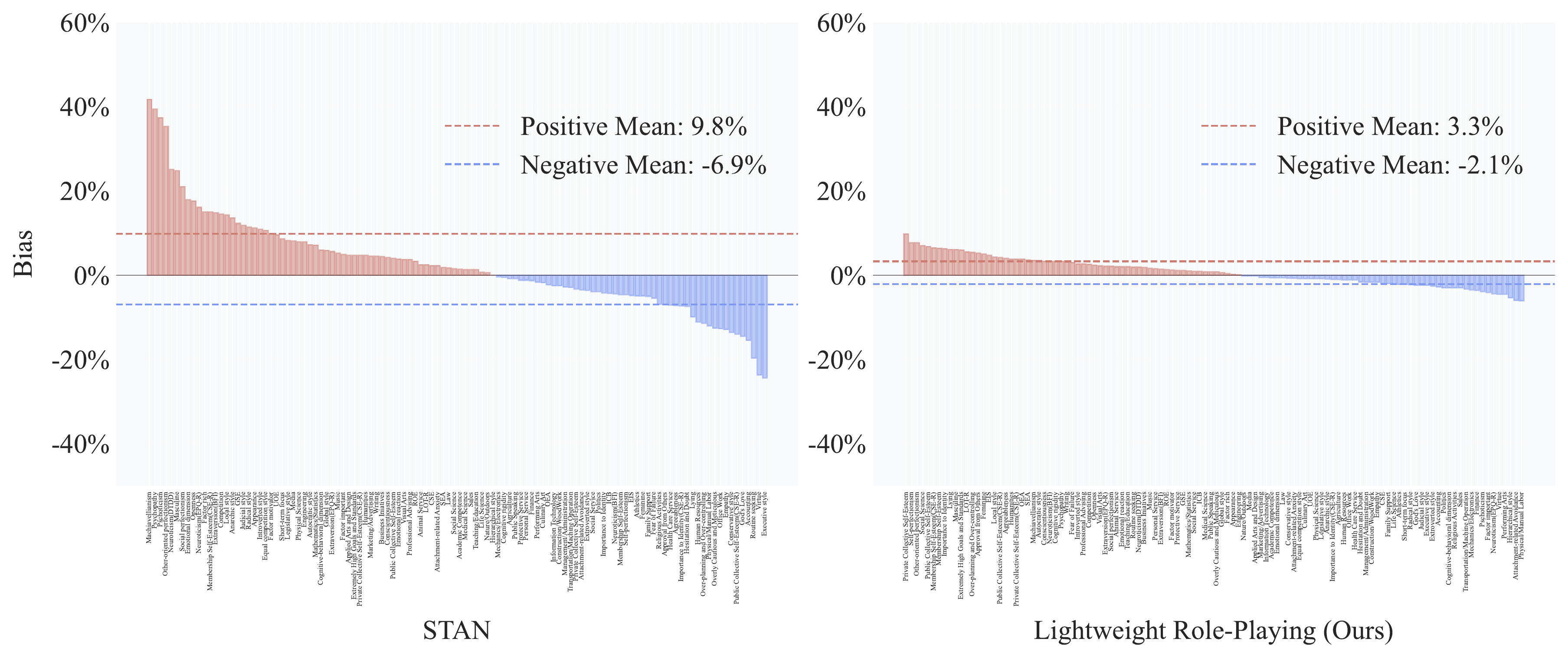}}
\caption[Caption for the list of figures]{
Psychometric Biases Introduced by Role-Playing Methods Designed to Bypass the Alignment of LLM.
}
\label{Bias_result}
\end{figure*}

\textbf{Finding 2: Format-transcoding-based jailbreaking cannot increase response rates in LLM psychometry.} After applying Base64 encoding or Caesar cipher encryption to the original psychometric scales, the average response rates of LLM did not show improvement compared to baseline. This is because, without altering the semantics, LLMs still tend to provide neutral or objective answers. In addition, since many LLMs are not proficient in encoding and cryptographic calculations, this type of method has even led to lower effective response rates. 

\textbf{Finding 3: By incorporating role-playing-type prompt prefixes, response rates in LLM psychometry can be effectively elevated.} The average response rates for the STAN method and our method are 81.49\% and 90.40\%, respectively, both exceeding the baseline rate of 70.12\%. This suggests that the use of anthropomorphic prompts can effectively reduce the proportion of objective and neutral responses given by LLMs.

\textbf{Finding 4: Semantics-camouflage-based jailbreaking is not the optimal method for enhancing the response rate in LLM psychometry.} Our method exhibits an average response rate that is 8.91\% higher than that of the STAN method, which aims to bypass the safety alignment restrictions of LLM (i.e., jailbreaking). Our proposed lightweight role-playing prompt is not designed to elicit harmful output from LLMs. Therefore, it is less likely to trigger the LLMs' shielding mechanisms like STAN does, achieving the highest response rate.

\subsection{Psychometric Biases}
Although Finding 3 confirms that role-playing methods can effectively increase LLM response rates, the semantics of these additional prompts may also introduce extra psychometric biases. To evaluate these biases, we used the LLM scores on a total of 112 subcategories of all scales under the baseline condition (i.e., using the original scales without additional prompts) as a comparison. Subsequently, we evaluated the psychometric biases introduced by both the STAN method and our proposed method separately. The bias calculation process is illustrated in Formula \ref{equation-1}, where N represents the number of items within the corresponding subcategory that received valid responses from both the evaluated method and the baseline condition. The evaluation results are presented in Figure \ref{Bias_result}.

\begin{equation}
\label{equation-1}
Bias_{Method:Sub.}=\frac{\sum_1^N(Score_{Method:Sub.}-Score_{Baseline:Sub})}{N*Score_{Max:Sub.}}*100\%
\end{equation}

\textbf{Finding 5: Semantics-camouflage-based jailbreaking prompts introduce significant psychometric biases.} The psychometric biases introduced by the STAN method, both positively and negatively, are notably higher than those introduced by our lightweight role-playing method. This corresponds with our expectations, as the STAN method (as its name implies, Strive To Avoid Norms) is designed to bypass LLM's safeguards against harmful content, inevitably leading to the creation of stronger virtual personas and, consequently, introducing pronounced psychometric biases.

\textbf{Finding 6: The biases of the lightweight role-playing method fall within a reasonable margin of error.} The average psychometric biases of our method are 3.3\% for positive biases and 2.1\% for negative biases. Among the 112 subcategories, 86 exhibit a bias of less than 4\%. We consider this level of bias to be reasonable, given that even human test subjects experience fluctuations in psychometric scores each time. Despite setting the temperature of the LLMs to zero, this does not guarantee identical responses in every test iteration. Therefore,  scores obtained under the baseline conditions should only be considered as a standard for comparison, rather than as definitive correct answers.

\subsection{Deviations Caused by Languages}

We conducted experiments using multiple languages for the same scales to evaluate the impact of the linguistic environment on the LLM psychometrics. Using the English scale scores as a compaired standard, we calculated the relative biases (deviations) of the scores obtained from measurements in other languages. The caculation process is as detailed in Formula \ref{equation-2}, where M represents the number of items within the corresponding scale subcategory that received valid responses in both the tested language and English.

\begin{equation}
\label{equation-2}
Bias_{Lang.:Sub.}=\frac{\sum_1^M(Score_{Lang.:Sub.}-Score_{EN:Sub})}{M*Score_{Max:Sub.}}*100\%
\end{equation}

\begin{figure}[H]
\centering
\includegraphics[width=0.5\textwidth]{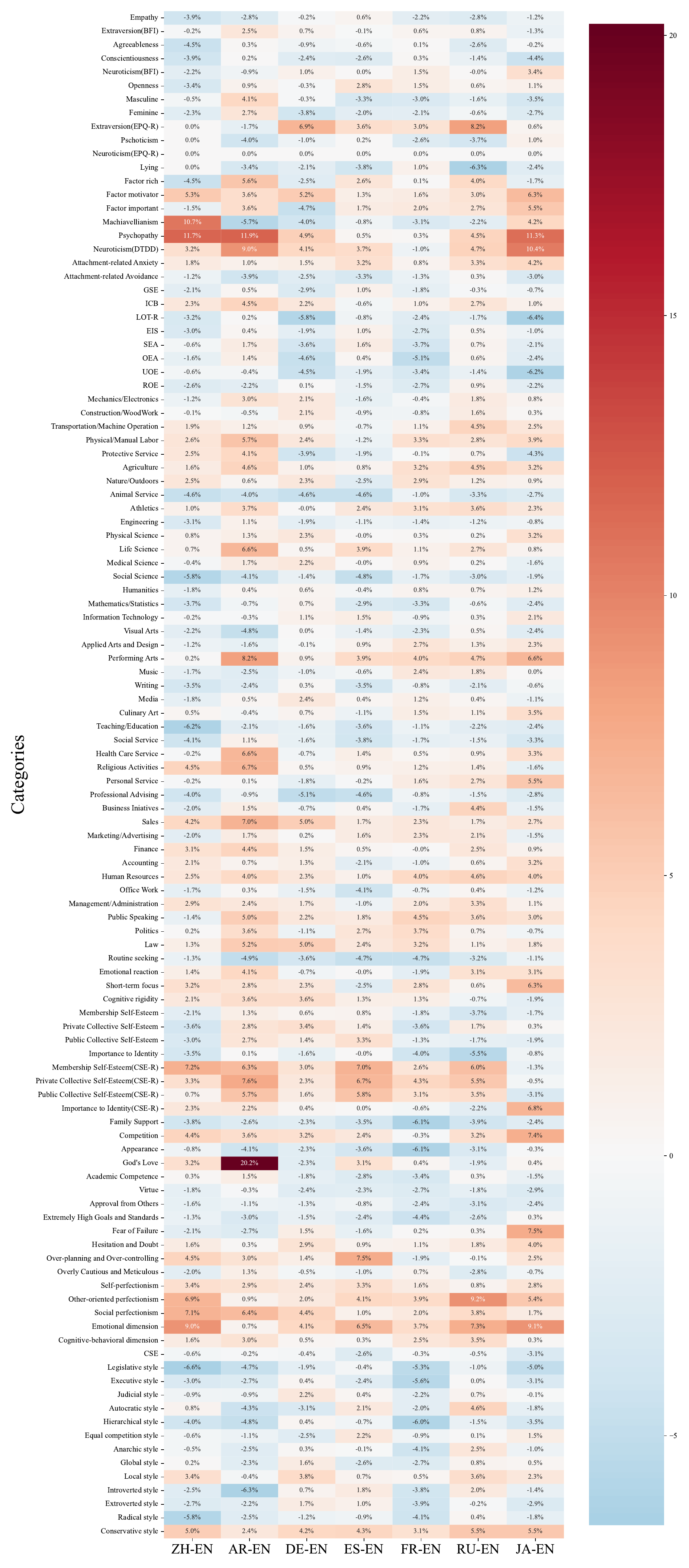}

\caption[Caption for the list of figures]{The psychometric scores of LLMs vary significantly when using different languages.
}
\label{heatmap}
\end{figure}

\textbf{Finding 7: LLMs indeed demonstrate significant psychometric deviations in various linguistic contexts.} As depicted in Figure \ref{heatmap}, typical examples of these deviations include, but are not limited to, the following:
\begin{itemize}

    \item (1) In the Arabic test, the scores of the \textit{God's Love} subcategory in the \textit{Contingencies of Self-Worth Scale (CSWS)} exhibited the most significant discrepancy compared to those of the English test, with a mean positive deviation of 20.2\%. Additionally, the \textit{Religious Activities} subcategory in the \textit{Comprehensive Assessment of Basic Interests (CABIN)}  showed an average positive deviation of 6.7\%.
    \item (2) In the \textit{Emotional Dimension} of the \textit{Multidimensional Perfectionism Scale (MPS)}, compared to the English test scores, the Japanese and Chinese test scores showed positive deviations of 9.1\% and 9.0\%, respectively. In addition, in the \textit{Fear of Failure} subcategory of \textit{Zi's Negative Perfectionism Questionnaire (ZNPQ)}, the Japanese respondents exhibited a 7.5\% positive deviation.
    \item (3) In the \textit{
Eysenck Personality Questionnaire-Revised (EPQ - R)}, the \textit{Extraversion} dimension of the personality scale in the Russian test shows a substantial positive deviation, averaging 8.2\%, compared to those of the English test.
    \item (4) In the French version of the \textit{Thinking Styles Inventory (TSI)}, the scores of \textit{Executive Style} and \textit{Hierarchical Style} exhibit significant negative deviations, with respective averages of 5.6\% and 6.0\%.
    \item (5) In the \textit{Life Orientation Test - Revised (LOT - R)}, which measures individual levels of optimism versus pessimism, a substantial negative deviation is observed in the Japanese test, with an average of 6.4\%. The German test exhibits the second highest negative deviation, averaging 5.8\%. 

\end{itemize}

\section{Discussion and Future Works}
\textbf{Is it possible to eliminate the above biases in LLM psychometrics?} Prefix prompts designed to enhance the response rate inevitably introduce additional biases in semantics. Future research efforts may propose more effective strategies to strike the balance between optimizing the response rate and minimizing additional biases. However, biases (deviations) arising from linguistic variations are not necessary to eliminate, given that LLMs are trained in multilingual corpora. Valuable future research should focus on developing more precise and comprehensive methodologies for measuring and interpreting these deviations.

\section{Conclusion}
To address the psychometric differences between LLMs and humans, we propose AIPsychoBench, a specialized benchmark to evaluate the psychological properties of LLMs. AIPsychoBench uses lightweight role-playing prompts to improve the effective response rate without introducing substantial psychometric biases. Experiments based on AIPsychoBench have yielded the first comprehensive and quantitative evidence of the linguistic impact on LLM psychometrics. Our work seeks to fortify the research foundation of machine psychology and to advance the interpretability of LLMs.

\balance

\printbibliography

\end{document}